\documentclass{article}

\usepackage{PRIMEarxiv}

\usepackage[utf8]{inputenc} % allow utf-8 input
\usepackage[T1]{fontenc}    % use 8-bit T1 fonts
\usepackage{hyperref}       % hyperlinks
\usepackage{url}            % simple URL typesetting
\usepackage{booktabs}       % professional-quality tables
\usepackage{amsfonts}       % blackboard math symbols
\usepackage{nicefrac}       % compact symbols for 1/2, etc.
\usepackage{microtype}      % microtypography
\usepackage{lipsum}
\usepackage{fancyhdr}       % header
\usepackage{graphicx}       % graphics
\graphicspath{{media/}}     % organize your images and other figures under media/ folder
\usepackage{pgffor}
\usepackage{cite}

\usepackage{xcolor}
\usepackage{amsmath,amssymb}
\usepackage{bm}
\usepackage{longtable}
\usepackage{caption}
\usepackage{subcaption}
\newcommand{\rr}{{\bm{r}}}

\newcommand{\md}{{\rm{d}}}
\newcommand{\me}{{\rm{e}}}
\newcommand{\mi}{{\rm{i}}}
\newcommand{\lacenet}{\textsc{LaceNET} }
\newcommand{\lacenetv}{\textsc{LaceNET}}
\newcommand*\samethanks[1][\value{footnote}]{\footnotemark[#1]}
\title{Realistic mask generation for matter-wave lithography via machine learning
}

\author{
Johannes Fiedler\thanks{The two first authors contributed equally to this work}\\
Department of Physics and Technology\\University of Bergen,
All\'eegaten 55\\5007 Bergen, Norway\\\textit{johannes.fiedler@uib.no}\\
\And 
Adri\`a Salvador Palau\samethanks\\
Department of Physics and Technology\\University of Bergen,
All\'eegaten 55\\5007 Bergen, Norway\\
\And
Eivind Kristen Osestad\\
Department of Physics and Technology\\University of Bergen,
All\'eegaten 55\\5007 Bergen, Norway\\
\And 
Pekka Parviainen\\Department of Informatics\\University of Bergen, HIB -
Thormøhlens gate 55\\ 5006 Bergen, Norway
\\
\And
Bodil Holst\\Department of Physics and Technology\\University of Bergen,
All\'eegaten 55\\5007 Bergen, Norway\\
}

\begin{document}
\maketitle

\begin{abstract}
Fast production of large area patterns with nanometre resolution is crucial for the established semiconductor industry and for enabling industrial scale production of next generation quantum devices. Metastable atom lithography with binary holography masks has been suggested as a higher resolution/low cost alternative to the current state of the art: extreme ultra violet (EUV) lithography. However, it was recently shown that the interaction of the metastable atoms with the mask material (SiN) leads to a strong perturbation of the wavefront, not included in existing mask generation theory, which is based on classical scalar waves. This means that the inverse problem  (creating a mask based on a desired pattern) cannot be solved analytically even in 1D. Here we present a machine learning approach to mask generation targeted for metastable atoms. Our algorithm uses a combination of genetic optimisation and deep learning to obtain the mask. A novel deep neural architecture is trained to produce an initial approximation of the mask. This approximation is then used to generate the initial population of the genetic optimisation algorithm that is able to converge to arbitrary precision. We demonstrate generation of arbitrary 1D patterns for system dimensions within the Fraunhofer approximation limit.
\end{abstract}

% keywords can be removed

\section{Introduction}\label{sec1}

All semiconductor device fabrication is currently based on pattern generation using mask based photolithography. The standing aim is to create patterns with smaller and smaller features at higher and higher information densities. The state of the art is EUV photolithography, which, with photons (light-waves) of a wavelength of 13.5 nm, should be able to produce patterns with a resolution of 6.75 nm according to the Abbe criterion. Smaller features could potentially be achieved using immersion and/or over-exposure or under-development, as is done with the 193-nm lightsource, EUVs predecessor. However, the problem is that in EUV lithography, due to the high energy of the photons  (91.8 eV), the pattern generation process is mediated by photo-generated secondary electrons, which can travel for several nm before inducing a reaction. Current experiments and theory indicate that the secondary electron blur radius for EUV is around 3 nm, which would limit the feature size that can be achieved to around 6 nm~\cite{Robinson}. This means that devices based on small quantum dots and individual atoms and molecules cannot be produced with EUV. Moving to wavelengths even
shorter than 13.5 nm, would just exacerbate the secondary electron issue - an alternative to photons is needed.

Already more than three decades ago lithography with atoms (matter-waves) was suggested as an alternative to photolithography. The reason for this is that for a given wavelength, the energy of an atom is much less than the energy of a photon.  A helium atom with the same wavelength as an EUV photon has a kinetic energy of only $E = h^2/(2m\lambda^2) \approx 0.011 \,\rm{eV}$, where $h$ denotes Planck’s constant and $m$ is the mass of the helium atom. This is almost a factor 10.000 less than the EUV photon. Some of the first atom lithography experiments created the patterns by depositing atoms directly on a substrate.  However, this method strongly limits the application areas, because it put restrictions on the materials which the pattern can be made of. This can be avoided by using metastable  atoms. When a metastable atom hits the substrate, it decays, and the decay energy is transferred to the substrate~\cite{Berggren1995,Baldwin2005,Ueberholz2002}. In 1995, Berggren \textit{et al.}~\cite{Berggren1995} demonstrated pattern generation in a thiol-based resist using a beam of metastable argon atoms manipulated by a light-field mask.  The energy released when a metastable atom decays is about 10 eV for argon and 20 eV for helium~\cite{Baldwin2005,Colli54}. A range of experiments have been done generating patterns by manipulating  atomic beams either by light or electrostatic fields~\cite{Berggren1995,Gardner2017,ADAMS1994143,Hinderthuer1998,Nesse2019}. However, these methods cannot be used to generate complex patterns with high resolution.  For high resolution pattern generation experiments have  been done focusing atom beams with lenses~\cite{Eder2017,Koch2008,Doak1999,Carnal1991,Eder2015,Eder2012,Patton2006,Barr2016,Reisinger2008,Fiedler_2017}, which were then used for serial writing of arbitrary patterns. However, for mass-scale production serial writing is not a suitable method.

The particular challenge for pattern generations with metastable atoms is that they decay when they impinge on a surface, so it is not possible to use masks made on substrates as is done in photolithography. The pattern must be generated by open areas in the substrate, where the beam can go through. This means any pattern with a closed path (i.e. a circle) would lead to a segment of the mask falling out.

In 1996, Fujita \textit{et al.}~\cite{Fujita1996} came up with an idea to circumvent this problem. They demonstrated pattern generation with metastable neon atoms, using a solid mask consisting of a distribution of uniformly sized holes (30 nm in diameter), etched into a silicon nitride membrane. The hole distribution was calculated using the theory of grid-based binary holography developed by Lohmann and Paris~\cite{Lohmann:67}, and later by Onoe and Kaneko~\cite{1979JElCo..62..118O} for scalar waves: This theory imposes the limitation  that the openings are all of the same size and positioned on a regular grid structure. The hole distribution is an approximated Fourier transform of the final, desired pattern. In recent publications binary philography was further explored, investigating issues such as  how many holes are needed to generate a particular pattern and what maximum resolution can be achieved with typical matter-wave wavelength scales~\cite{Nesse2017,Nesse2019}. However, earlier this year it was shown that the wavefront of metastable helium atoms is perturbed by dispersion forces when it passes through a hole in a silicon nitride membrane~\cite{Johannes21}, an effect not considered in the previous publications. Dispersion forces, in this case, the Casimir--Polder forces, are caused by the quantum-mechanical ground-state fluctuations of the electromagnetic field in the absence of charges~\cite{Scheel2008}. Due to the field fluctuations, the helium atom will be polarised for a short amount of time. The resulting induced dipole moment then interacts with the dielectric membrane via dipole-dipole interactions. These forces decay dramatically with the distance, $r^{-3}$-power law~\cite{PhysRevA.85.042513}, but play a significant role on the nanometre length scale and has been observed in several matter-wave diffraction experiments~\cite{Doak1999,Arndt1999,Brand2015,Hemmerich2016,PhysRevLett.125.050401}.

For Fujita \textit{et al.}, the mask holes were so big that the dispersion force interaction was negligible. Hence, they could use the standard binary holography theory to generate the mask design for their patterns. However, for smaller mask holes, which are required for high resolution, the effect of the dispersion force interaction becomes significant. In Ref.~\cite{Johannes21}, it is shown that metastable helium atoms cannot penetrate holes in 5 nm thick silicon nitride membranes with less than 2 nm diameter. 

The aim of this paper is to establish a theoretical framework for mask-based matter-wave lithography, initially in one dimension, so that for a given desired pattern, atom wavelength and dispersion force interaction a mask can be generated.  Mathematically this is a so called inverse problem~\cite{inverse}. We now explain why machine learning is the only realistic approach to solving this problem even in the one dimensional case:

As discussed above, existing binary holography theory for scalar waves, uses a grid of holes with equal size as the base for the masks and assumes system dimensions and wavelength that fulfil the well known optics condition: Fraunhofer approximation~\cite{BornWolf:1999:Book}. In the Fraunhofer approximation the inverse problem is reduced to a Fourier transform of the desired pattern, which yields the required mask. The mask needs to be sampled according to the Nyquist--Shannon sampling theorem to account for the finite size of the  holes and mask. In one dimension, this can be done analytically, see Sec.~\ref{sec:em}. However, this solution  cannot be adapted to matter waves, because the dispersion force interaction between the particles and the diffraction object induces a complex phase distribution of the matter wave~\cite{Fiedler_2017}.

An alternative approach to Fourier transformation, starts from convolution of the mask function with the so-called point-spread functions~\cite{BRAAT2008349} representing a basis of elementary openings. This approach however rises several issues: (i) the complex structure of the required integral kernel (analogously to the Airy disc for matter waves~\cite{Hemmerich2016}) which (ii) does not scale linearly with the dimension of the holes~\cite{Johannes21} and (iii) would require additional optimisation algorithms to consider a variation of the hole sizes.

\begin{figure}
    \centering
    \includegraphics[width=0.7\linewidth]{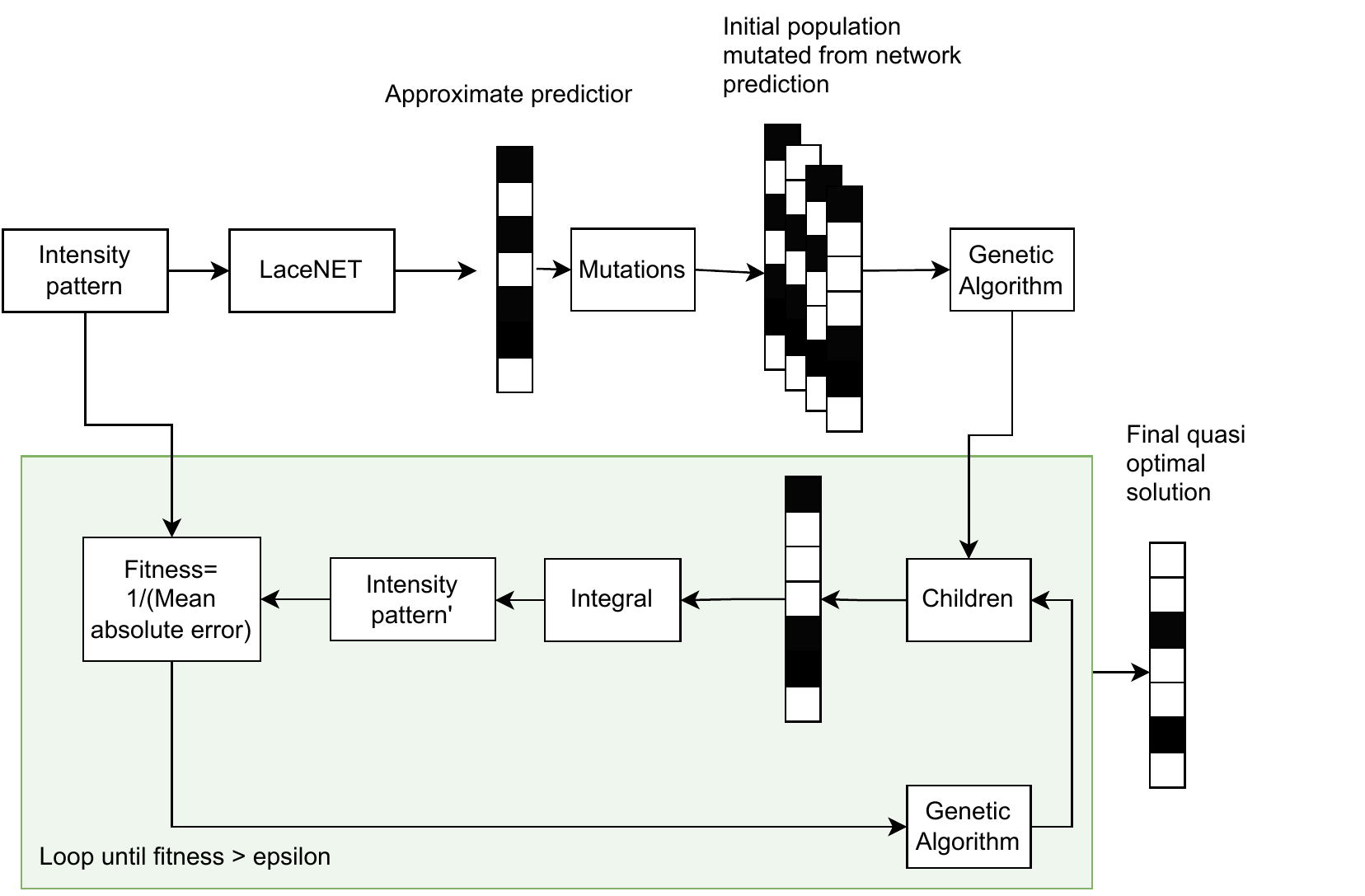}
    \caption{\label{fig:General_framework} General mask generation framework used in this paper. First, the desired intensity pattern is fed to \lacenetv. \lacenet then computes an approximated mask that is expected to produce this pattern. From this mask, an initial population is generated and fed to a genetic algorithm that is run until the accuracy condition is met. }
\end{figure}

In this paper, we use machine learning as a new approach to mask generation for metastable atoms, in particular we include the dispersion force interaction and opens up for the possibility that holes can be of different sizes and not positioned on a regular grid structure. Our machine learning architecture is based on a combination of deep learning and genetic algorithms. The problem is solved in two steps: first, a deep neural network~\cite{goodfellow2016deep} is trained on large amounts of generated data. Second, the neural network resolves the desired diffraction pattern, producing a mask. This mask is then mutated to produce an initial population that is fed to a genetic algorithm~\cite{herrera1996genetic}. The genetic algorithm then is able to quickly converge to a mask that produces the desired diffraction pattern with arbitrary precision (see Fig. \ref{fig:General_framework}).

Deep learning has been used extensively to solve inverse problems due to neural networks being universal function approximators~\cite{hornik1989multilayer}. For example, it has been applied to light inverse scattering problems~\cite{chen2020review}. The high-level combination used in this paper (deep learning plus genetics algorithm) has been used in photonic device design~\cite{ren2021genetic}, inverse molecular design~\cite{nigam2021janus}, and robot manipulators~\cite{koker2013genetic}, among others. However, to our knowledge, it has never been used to solve an inverse diffraction problem, where a diffraction mask is recovered from a diffraction pattern.

\section{Results}\label{sec11}

\subsection{Mask generation framework}
The general idea of our mask generation framework follows: first, a deep convolutional neural network is used to approximate a general solution to the inverse problem. This neural network is trained with a large data set of randomly generated examples that map a mask to a diffraction pattern (see Sec.~\ref{sec:gen_training_data} for details of how the examples are generated). Masks are represented as binary sequences of finite length where 1 represents openings in the mask and 0 blocked space.

Once the approximate solution is obtained, this solution (a binary mask) is randomly mutated. The random mutations of this approximate solution form the initial population of a genetic algorithm. The genetic algorithm --- known to be an efficient optimiser for binary sequences~\cite{zhao2009optimisation} --- further refines the mask provided by the neural network until a convergence constraint is satisfied.

\subsubsection{\lacenetv: a neural network for approximate inversion of the Kirchhoff diffraction formula}
\lacenet is a deep convolutional neural network that takes as features (inputs) the desired intensity pattern and its fast Fourier transform and has to output the mask pattern, a binary intensity map. We choose the real-input fast Fourier transform as one of the network's features because the Fourier transform of the diffraction pattern plays a significant role in far-field diffraction~\cite{BornWolf:1999:Book,Nesse2019}. Thus, the diffraction patterns have periodic components that can be efficiently compressed by using its Fourier decomposition, and for this reason, allowing for better learning by our neural network. 

We split the network graph in three towers, each specialised in one of the three different features (the intensity map, and the real and intensity parts of the fast Fourier transform). Within each tower, we use skip connections - a mechanism that has been used effectively in the U-Net architecture, an inverse architecture for light scattering problems~\cite{wei2018deep}. Skip connections simply consist of taking the input to a neural network layer (or succession of layers) and summing it after the neural network layer has been applied to it~\cite{orhan2017skip}.

\begin{figure}
    \centering
    \includegraphics[width=0.6\linewidth]{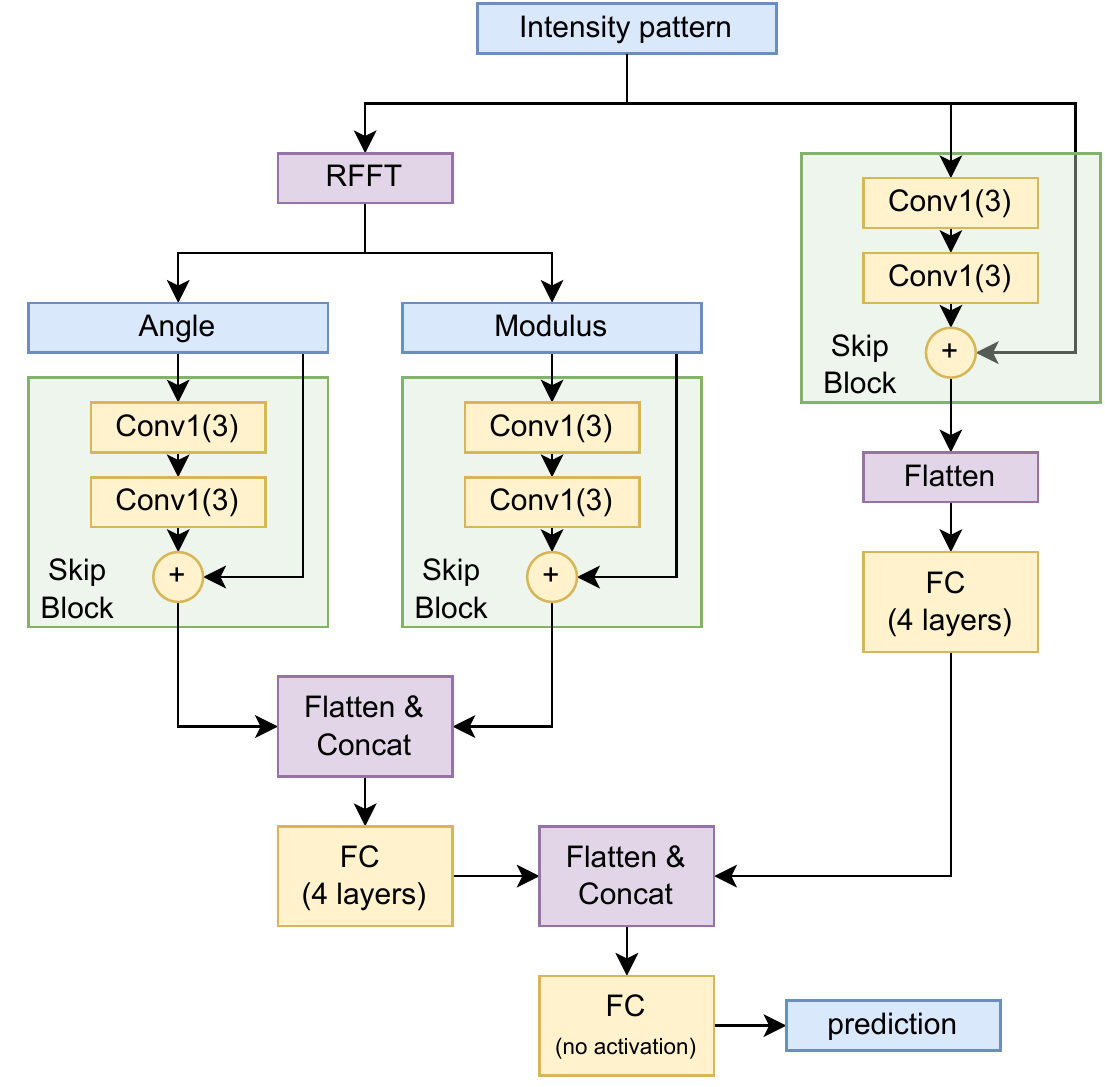}
    \caption{\label{fig:lacenet} \lacenet architecture. A three-tower architecture . The two towers in the left are 1-D convolutional layers with skip blocks that ingest the angle and the modulus of the right fast Fourier transform of the input. The tower in the right is also composed by one skip block with two convolutions. Fully connected layers and concatenations are used to combine the information coming from the three blocks.}
\end{figure}

Finally, we use a fully connected layer to learn the predicted mask (see fig.~\ref{fig:lacenet} for the full architecture). Note that unlike many popular computer vision architectures, we do not use batch normalisation due to the reason that after extensive experimentation, we did not see any benefit for our task.
We train the neural network with the focal loss. A loss commonly used to train segmentation masks in computer vision~\cite{jadon2020survey}. The focal loss is derived from the binary cross entropy loss $l_{\rm BC}$:
\begin{equation}
    l_{\rm BC}(y,\hat{y})=\left[y\log\hat{y}+(1-y)\log(1-\hat{y})\right] \,.
\end{equation}
By summing over the all mask points with the total number $N_{\rm points}$, we reach the total loss per mask $L$:
\begin{equation}\label{eq:BCE_loss}
    L(y,\hat{y})=-\frac{1}{N_{\rm pixels}}\sum_{\rm mask}l_{\rm BC}(y,\hat{y}) \,,
\end{equation}
with $y$ is the true value of each pixel in the mask sequence and $\hat{y}$ is the value predicted by the neural network.

The focal loss is a variation of the binary cross-entropy loss and reads
\begin{equation}\label{eq:focal_loss}
    l_{\rm FO}^i(y,\hat{y})=-\alpha_i\left[1-\mathrm e^{l_{\rm BC}(y,\hat{y})}\right]^\gamma l_{\rm BC}(y,\hat{y}) \,.
\end{equation}
This loss function is designed to achieve two purposes: (I) weight the classes (in our case 1, and 0) according to their rarity through the weights $\alpha_i$, so that they contribute equally to the loss, and (II) reduce the loss of easily-classified examples, so that the network can focus on the harder parts of the mask.

\lacenet can be defined as a deterministic non-linear function that depends on its parameters $w$. The optimisation problem that \lacenet solves as training is to find the set of parameters $w$ that minimise the loss function over the training set $\Psi$,
\begin{equation}
    w^*=\min_w\left[\sum_{\Psi} L(y,\hat{y})\right]\,.
\end{equation}
\lacenet is trained on a dataset $\mathcal{D}$ of 300k samples, randomly generated as described in Sec.~\ref{sec:gen_training_data}. As it's standard in deep learning, this dataset is split in three smaller subsets: a training set $\mathcal{D}_{\rm train}$, a test set $\mathcal{D}_{\rm test}$, and a validation set $\mathcal{D}_{\rm val}$. The split used is 81/9/10 (train/validation/test). One tenth of the data is reserved for test and one tenth of the remaining training data is reserved to validate hyper-parameters. The training of \lacenet is split in two parts: hyperparameter tuning (which optimises for hyperparameters of the neural network such as learning rate, loss function type etc), and the training of the network itself (i.e. the training of its weights and biases).

We use a variation of Bayesian search to train the hyperparameters of the neural network, and evaluate the validation dataset using Mean Squared Error over the result of the integrals of the produced mask. We choose to evaluate using Mean Squared Error because that is the metric that really tells whether the mask produces the required pattern. In the discretised space, the inverse may not be unique and small differences in the binary sequence that represents the mask might have a strong effect on the focal loss, while actually affecting the mean squared error loss of the integrated mask to a much lesser extent.

\begin{figure}
    \centering
    \includegraphics[width=0.7\linewidth]{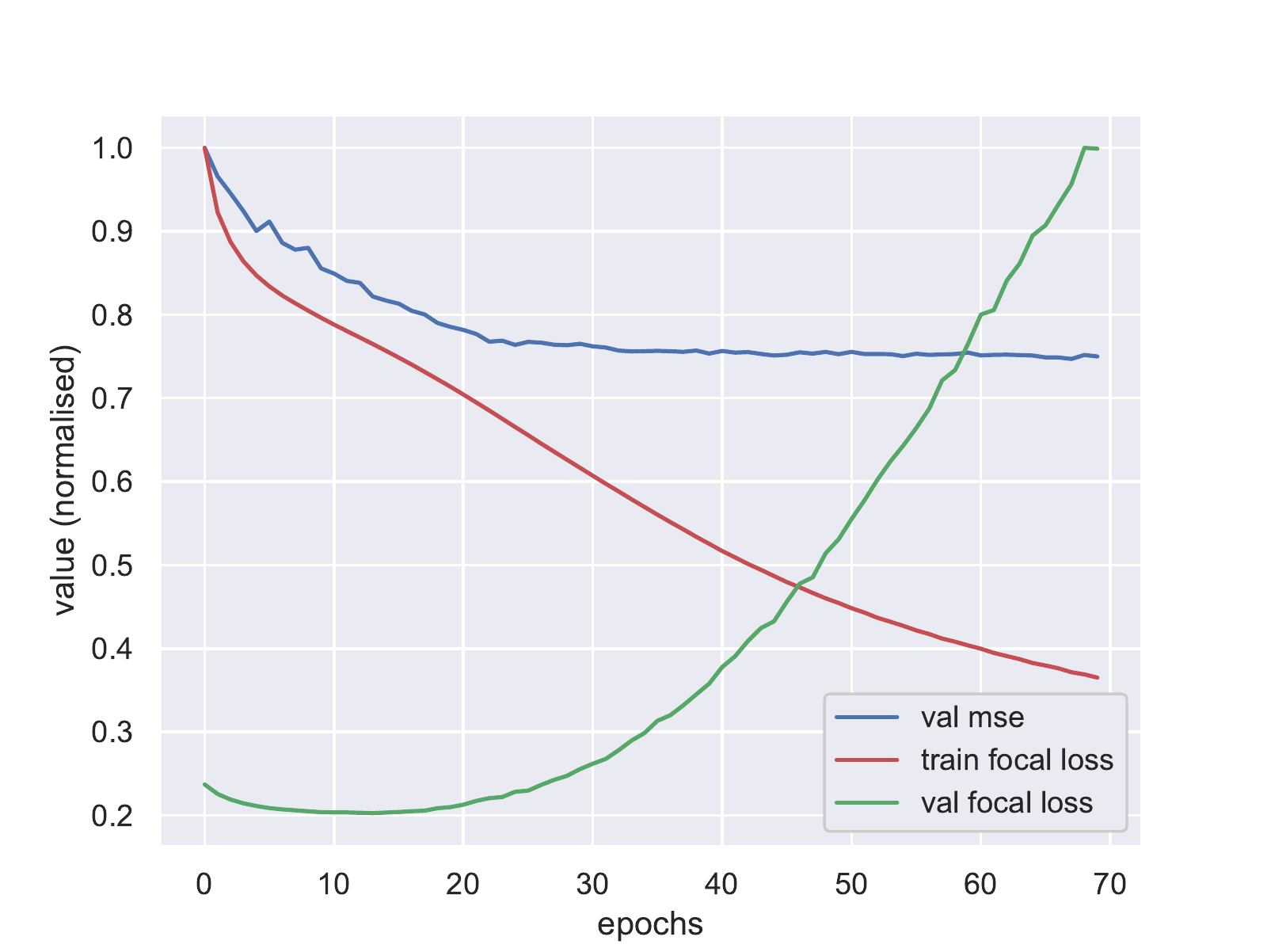}
    \caption{\label{fig:lacenet_training} Training (red) and validation (green) focal loss (eq. \ref{eq:focal_loss}) compared with mean squared error computed over the validation set after performing the integral of the solution (blue). Note how the focal loss over the validation seems to exhibit a clear over-fitting effect. However, when calculating the mean squared error of intensity produced by the mask, the network is actually seen to improve gradually as the epochs advance. This is likely due to two effects (i) the intensity integral does not have a unique inverse when discretised, (ii) after a certain number of epochs, the masks produced become very stable as the 0.5 thresholds convert the continuous output of the network into a discrete binary mask.}
\end{figure}

For the training of the neural network, we use the best optimisation algorithm (typically Adam \cite{kingma2014adam} or RMSProp \cite{hinton2012neural}) that is returned by the hyperparameter search. Training is stopped using early stopping with patience 10 and delta 0.001. Full details of this procedure and the space over which hyperparameter search is performed can be found in the supplementary information. The hyperparameters used to produce the results presented here are optimiser: Adam, learning rate: 0.00016, batch size: 225, $\alpha$: 0.439, $\gamma$: 5.952.
Figure~\ref{fig:lacenet_training} shows the focal loss and the mean squared error after integrating the masks evaluated over the validation dataset.

\subsubsection{Genetic algorithm}
Genetic algorithms are stochastic classical evolutionary algorithms, that is algorithms that dynamically change in order to optimise a fitness function $F$. Genetic algorithms are inspired by Darwin's theory of evolution that describes how the genes in the population evolve according to their capacity of reproducing and mutating. In genetic algorithms, the population is not an actual population of living beings but instead a set of solutions for a particular optimisation problem (known as chromosomes). Furthermore, mutation is not necessarily the result of natural stochastic processes but occurs according to different rules that can be decided by the programmer.  For each chromosome in the population a fitness value can be calculated using a pre-defined formula set at the choice of the researcher. Finally, the reproductive process of life is simulated by combining different chromosomes (parents) into next generations of offspring. The reproductive process can also be defined by the researcher \cite{eiben2003introduction}.

Genetic algorithms are very robust optimisation algorithms that are especially suited to work with sequences of categorical variables (such as the four types of bases present in DNA: adenine (A), cytosine (C), guanine (G), and thymine (T)). The masks that we want to obtain in this paper are a perfect example of that: they are fully defined by a sequence of binary values.

The genetic algorithm that we use here is applied to the solution by \lacenet, the fitness function is simply the inverse of the mean absolute error of the generated diffraction pattern plus a small numerical constant $\alpha$.

\begin{equation}
    F=\frac{1}{\alpha+\sum_{r_i}  \lvert\tilde{P}(r_i)-\psi(r_i;{\bm{x}},{\bm{d}})\psi^\star(r_i;{\bm{x}},{\bm{d}})\lvert}.
\end{equation}
$\tilde{P}(r_i)$ here is the absolute square value of the  wave function discretised over a grid of radial coordinates $r_i$ measuring the distance to the centre of the diffraction pattern. $\psi(r_i;{\bm{x}},{\bm{d}})$ is the wave function that is made to depend on the the positions ${\bm{x}}=\left(x_1,x_2,\dots\right)$ and thicknesses ${\bm{d}}=(d_1,d_2,\dots)$ of the grating (mask) openings. A detailed description of how the wave function is computed can be found in Sec. \ref{sec:forward_prop}. To run our genetic algorithm we use a pygad: a well-known genetic algorithm solver for python \cite{gad2021pygad}. Within pygad we use the following hyper-parameters (obtained through grid search). (i) uniform crossover, in which offspring is formed by random recombination of parents' chromosomes. (ii) fit parent persistence: the fittest chromosomes are carried on to the next generation. (iii) initial population size of 50 chromosomes. (iv):  The number of solutions to be selected as parents is set to 7. 

\begin{figure}
\centering
\begin{minipage}{.49\textwidth}
  \centering
  \includegraphics[width=\linewidth]{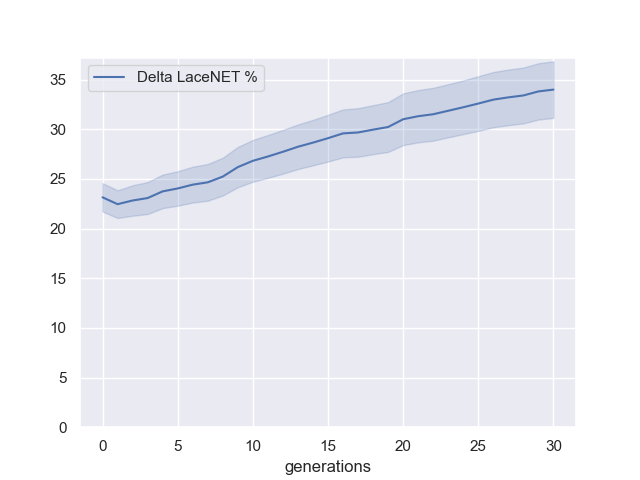}
  \captionof{figure}{Mean difference in percentage between using \lacenet to generate the initial population for the genetic algorithm and using a random approach. Results averaged over 100 never-seen randomly sampled test masks. Each mask is approximated by the generated algorithm 50 times. Error bars (shaded area) are 4$\sigma$. Note how \lacenet outperforms the randomly-generated population significantly, with the percentage difference growing larger as the genetic algorithm runs more generations.}
  \label{fig:deltapercentage}
\end{minipage}%
\hfill
\begin{minipage}{.49\textwidth}
  \centering
  \includegraphics[width=\linewidth]{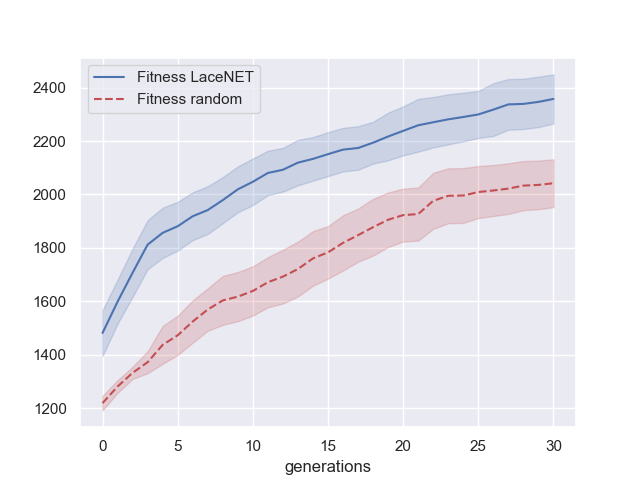}
  \captionof{figure}{Average fitnesses $F$ for an arbitrary mask after several generations of the genetic algorithm. The plot shows the fitness of the solution when the initial population is given by \lacenet (blue) compared with an initial population sampled randomly (red). Here shown an example where \lacenet helps converging to a much more precise solution. Error bars (shaded area) are 4$\sigma$, equivalent to 99.99\% certainty.}
  \label{fig:random_vs_lacenet}
\end{minipage}
\end{figure}

The initial population is not generated by pygad but by our own bespoke script. The script takes as an input the number of mutations allowed in each chromosome and generates a population by randomly mutating the output of \lacenet. \lacenet's solution is also kept as part of the initial population. The results presented here have 15 mutations in the chromosome. Mutations are assigned sampling randomly from an discrete uniform distribution - which means that a gen randomly set to mutate can maintain its initial value if it is sampled again from he distribution.

Figure~\ref{fig:deltapercentage} shows the average difference in percentage when comparing using \lacenet to generate the initial solution population and using a random initial population. Note that within $4\sigma$ error bars \lacenet outperforms a naive genetic approach. More importantly perhaps, \lacenet becomes better compared to using a randomly initiated genetic algorithm the more generations that the genetic algorithm is run, showing that the edge provided by the neural network increases with computational time.

\begin{figure}
    \centering
    \includegraphics[width=0.7\linewidth]{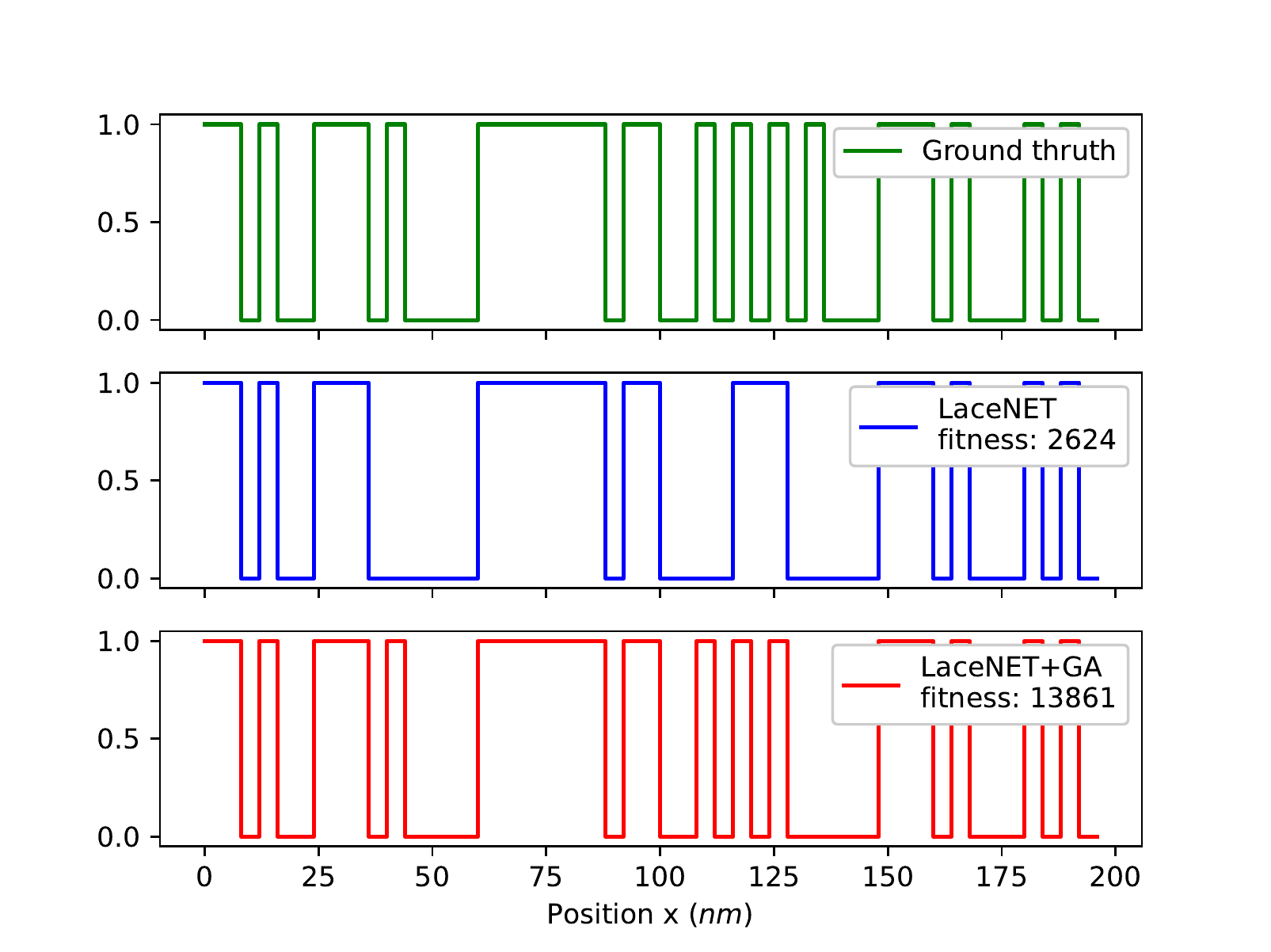}
    \caption{\label{fig:maskexample} Example of a mask returned by \lacenet (blue) and by \lacenet enhanced by the Genetic Algorithm after 2000 generations compared (red) to the ground truth (green). Note how the genetic algorithm introduces certain openings in the mask that were missed by the neural network, thus increasing the general fitness of the solution. Note as well how \lacenet (blue) provides a reasonable first approximation to the mask design.}
\end{figure}

Figure~\ref{fig:random_vs_lacenet} shows an example for a particular mask in which \lacenet performs significantly better than random. Results for all 100 randomly generated masks can be found in Appendix \ref{secA2}. As shown above, when all masks are averaged, \lacenet is found to outperform random significantly (by more $4\sigma$). Figure~\ref{fig:maskexample} compares the solutions returned by \lacenet and the genetic algorithm for a particular inverse problem. Note how \lacenet provides a reasonable first approximation to mask design and how further Genetic iteration incorporates small changes (mutations) that make the solution closer to the ground truth.
%\begin{figure}
%    \centering
%    \includegraphics[width=0.7\linewidth]{Delta_percentage_lacenet.png}
%    \caption{\label{fig:deltapercentage} Mean difference in percentage between using \lacenet to generate the initial population for the genetic algorithm and using a random approach. Results averaged over 100 never-seen randomly sampled test masks. Each mask is approximated by the generated algorithm 50 times. Error bars (shaded area) are 4$\sigma$. Note how \lacenet outperforms the randomly-generated population significantly, with the percentage difference growing larger as the genetic algorithm runs more generations.}
%\end{figure}

%\begin{figure}
%    \centering
%    \includegraphics[width=0.7\linewidth]{Figure_4sigma_onetraj.png}
%    \caption{\label{fig:random_vs_lacenet} Average fitnesses $F$ for an arbitrary mask after several generations of the genetic algorithm. The plot shows the fitness of the solution when the initial population is given by \lacenet (blue) compared with an initial population sampled randomly (red). Here shown an example where \lacenet helps converging to a much more precise solution. Error bars (shaded area) are 4$\sigma$, equivalent to 99.99\% certainty.}
%\end{figure}

\subsection{Forward propagation}\label{sec:forward_prop}
Here we describe how the training patterns are generated by forward propagation of matter-waves through randomly generated masks.  The diffraction at a dielectric interface can be described via classical waves due to the duality of waves and particle~\cite{messiah2014quantum}, which can be derived via Kirchhoff's diffraction formula~\cite{BornWolf:1999:Book}. This formula determines the propagation of a classical wave with amplitude $a_0$
\begin{eqnarray}
\psi_{\rm{cl}}({\bm{r}}_{\rm D}) = \frac{a_0 k_0}{2\pi \mi}\int \md^2 r_{\rm{i}}\, T({\bm{r}}_{\rm i}) \frac{\me^{\mi k_0\left(r_{\rm S,i}+r_{\rm i,D}\right)}}{r_{\rm S,i}r_{\rm i,D}} \frac{\cos\vartheta+\cos\vartheta'}{2}\,, \label{eq:Kirchhoff}
\end{eqnarray}
through an interface, in analogy to Huygens--Fresnel principle~\cite{Landau}. The interface is usually described by a transmission function $T$, which varies (for electromagnetic waves) between $1$ for transmission and $0$ for absorption. The source of the wave is located at $\rr_{\rm S}$ (subscript S for source), and the grating  is placed at the intermediate points $\rr_{\rm i}$ (subscript i for intermediate), leading to the relative coordinates $\rr_{\rm S,i}={\bm{r}}_{\rm S} - {\bm{r}}_{\rm i}$ to the grating and continues the propagation to the detector $\rr_{\rm D}$ (subscript D for detector) $\rr_{\rm i,D} =\rr_{\rm i}-\rr_{\rm D}$. The wave vector is related via the de-Broglie wavelength $k_0 = 2\pi/\lambda_{\rm dB} = p/\hbar$ with the particle's momentum ${\bm{p}}= p {\bm{e}}_p$.~\cite{messiah2014quantum} The geometric correction angles $\vartheta$ and $\vartheta'$ are the angles between the aperture's normal ${\bm{n}}$ and the wave's propagation directions, $\rr_{\rm S,i}$ and $\rr_{\rm i,D}$, respectively. 

By considering plane waves passing the obstacle, the propagation lengths are dominated by the distances between the source and the interface $r_{\rm S,i}\approx L_1$ and between the interface and the detector $r_{\rm i,D} \approx L_2$. This approach simplifies Kirchhoff's diffraction formula~(\ref{eq:Kirchhoff}) to the Fourier transform of the transmission function and is known as Fraunhofer approximation
\begin{eqnarray}
\psi_{\rm{cl}}(r) = \frac{a_0 k_0}{2\pi\mi}\frac{\me^{\mi k_0 \left[L_1 + L_2 + r^2/(2L_2)\right]}}{L_1L_2}\int\md A\, T \me^{\mi \frac{k_0}{L_2}{\bm{r}}\cdot{\bm{s}}}\,,
\end{eqnarray}
where $r$ denotes the position at the screen. Furthermore, the transmission function is invariant along the direction of the slit and, thus, the wave at the screen is just given by the one-dimensional Fourier transform of the transmission function
\begin{eqnarray}
\psi_{\rm{cl}}(r) = \frac{a_0 k_0}{2\pi\mi}\frac{\me^{\mi k_0 \left[L_1 + L_2 + r^2/(2L_2)\right]}}{L_1L_2}\int\md x\, T(x) \me^{\mi \frac{k_0}{L_2}rx}\,. \label{eq:1d}
\end{eqnarray}
Assuming the mask to be an array of slit openings the total transmission function can be determined by the superposition of the single slits
\begin{equation}
    T(x) = \sum_n t(x-x_n;d_n)\,,\label{eq:transmission}
\end{equation}
with the centre coordinate of each opening $x_n$. We allow each opening to vary in its width $d_n$. By inserting the transmission function~(\ref{eq:transmission}) into the wave equation~(\ref{eq:1d}), the field at the screen decomposes into the superposition of single slits with a phase correction due to the corresponding spatial shift
\begin{eqnarray}
\psi(r) = \frac{a_0 k_0}{2\pi\mi}\frac{\me^{\mi k_0 \left[L_1 + L_2 + r^2/(2L_2)\right]}}{L_1L_2}\sum_n\me^{-\mi \frac{k_0}{L_2}rx_n}\int\md x\, t(x;d_n) \me^{\mi \frac{k_0}{L_2}rx}\,. \label{eq:wave}
\end{eqnarray}
The interference pattern is given by the absolute square value of the wave function
\begin{equation}
P(r) = \psi(r)\psi^\star(r) \,.
\end{equation}
This yields that a target pattern $\tilde{P}(r)$ can be approximated by minimising error function
\begin{eqnarray}
E({\bm{x}},{\bm{d}}) = \int \md r \left(\tilde{P}(r)-\psi(r;{\bm{x}},{\bm{d}})\psi^\star(r;{\bm{x}},{\bm{d}})\right)^2\,,\label{eq:error}
\end{eqnarray}
with respect to the positions ${\bm{x}}=\left(x_1,x_2,\dots\right)$ and thicknesses ${\bm{d}}=(d_1,d_2,\dots)$ of the grating openings. Ordinarily, one would apply a least-square fitting algorithm to reduce the error~(\ref{eq:error}) concerning the positions and grating openings. 
\subsubsection{Propagation of electromagnetic waves}\label{sec:em}
In the absence of dispersion forces, which is achieved for electromagnetic waves, the wave propagation~(\ref{eq:wave}) can be calculated explicitly leading to
\begin{equation}
  \psi(r) = \frac{a_0 }{\pi\mi}\frac{\me^{\mi k_0 \left(L_1 + L_2 + r^2/(2L_2)\right)}}{L_1 r}\sum_n\me^{-\mi \frac{k_0}{L_2}rx_n}\sin\left(\frac{k_0 r d_n}{2L_2}\right) \,, \label{eq:emw}
\end{equation}
and, thus, the interference pattern reads
\begin{equation}
P(r) =\frac{a_0^2}{\pi^2 L_1^2 r^2}\sum_{n,m} \me^{\mi \frac{k_0}{L_2}r\left(x_m-x_n\right)}\sin\left(\frac{k_0 r d_n}{2L_2}\right) \sin\left(\frac{k_0 r d_m}{2L_2}\right) \,.
\end{equation}
To analyse the relation between the interference pattern $P(r)$ and the transmission function parameters ${\bm{x}}$ and ${\bm{d}}$ concerning the resolution and contrast of the interference pattern, a Fourier analysis is required, which yields~\cite{kammler2000first}
\begin{eqnarray}
    \lefteqn{P(\kappa) = -\frac{a_0^2}{2 L_1^2} \sum_{n,m} \left\lbrace \left(\kappa-\frac{c_1}{2\pi}\right)\left[2\Theta\left(\kappa-\frac{c_1}{2\pi}\right)-1\right]+\left(\kappa-\frac{c_2}{2\pi}\right)\left[2\Theta\left(\kappa-\frac{c_2}{2\pi}\right)-1\right]\right.}\nonumber\\
    &&\left.
    +\left(\kappa-\frac{c_3}{2\pi}\right)\left[2\Theta\left(\kappa-\frac{c_3}{2\pi}\right)-1\right]+\left(\kappa-\frac{c_4}{2\pi}\right)\left[2\Theta\left(\kappa-\frac{c_4}{2\pi}\right)-1\right]
    \right\rbrace\,,
\end{eqnarray}
with the Heaviside function $\Theta$ and the frequency shifts
\begin{eqnarray}
    c_1= \frac{k_0}{2L_2}\left(2x_m-2x_n-d_n+d_m\right) \,,&& 
    c_2= -\frac{k_0}{2L_2}\left(-2x_m-2x_n-d_n+d_m\right) \,,\\
    c_3= \frac{k_0}{2L_2}\left(2x_m-2x_n+d_n+d_m\right) \,,&&
    c_4= -\frac{k_0}{2L_2}\left(-2x_m-2x_n+d_n+d_m\right) \,.
\end{eqnarray}
This analysis illustrates that the relevant contributions is located in the range $\kappa =\mathcal{O}\left(10^{-5} /\lambda_{\rm dB}\right)$ for typical parameters used in matter-wave lithography [spatial resolution of the mask $\Delta x =\mathcal{O}(\rm{nm})$ and distance between the mask and the object plane $L_2=\mathcal{O}(100\,\rm{\mu m})$]. By increasing the mask's extension to be in the same order as $L_2$, the entire spectral resolution will be in the range $10^{-5}/\lambda_{\rm dB} \le \kappa  \le 1/\lambda_{\rm dB}$.

However, due to the presence of dispersion forces, the application of such method would result in several numerical issues caused by the complexity of the dependence on these quantities, which will be illuminated in the next section.

\subsubsection{Dispersion  force interactions}
The dispersion force interaction between an atom and a dielectric membrane can be approximated by~\cite{Johannes21}
\begin{equation}
    U_{\rm CP,app}(\rr) = -\frac{9 C_3}{\pi}\int \frac{\md^3s}{\left[\left({\bm{s}}-\rr\right)\cdot \left({\bm{s}}-\rr\right)\right]^3}\,,\label{eq:Ham}
\end{equation}
with the $C_3$-coefficient denoting the interaction strength between the atom and a plane built of the same material as the membrane, the position of the atom  $\rr$, and the integration volume $\md^3s$ is bounded by the membrane's surface.

\begin{figure}
    \centering
    \includegraphics[width=0.7\columnwidth]{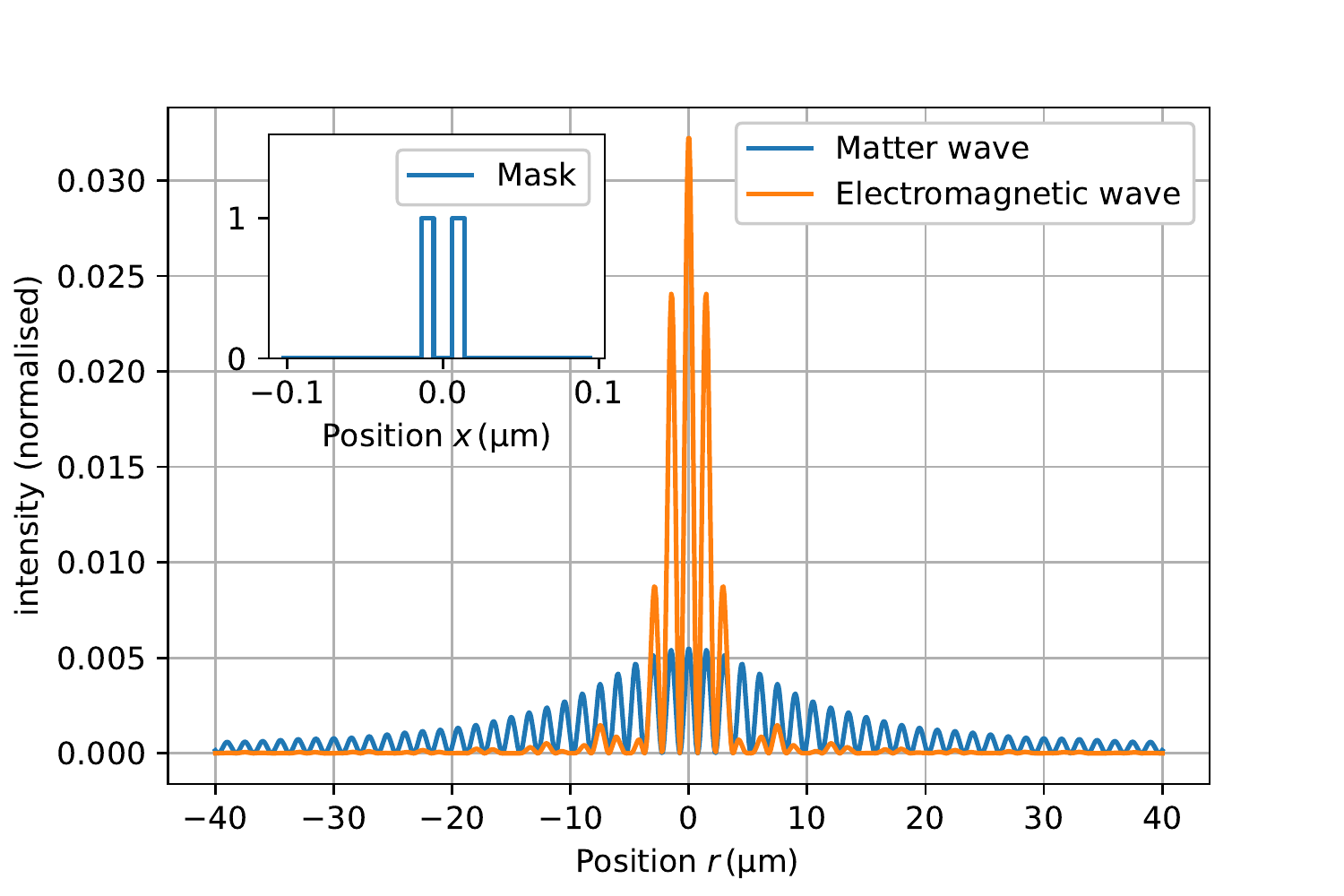}
    \caption{Comparison of the interference patterns obtain by passing a light wave (electromagnetic wave) (orange) and a metastable helium atom matter wave (blue) through the same SiN double-slit mask of 5 nm thickness with 8 nm openings separated by an 8 nm wall. The inset shows the mask used for both cases.}
    \label{fig:comparison}
\end{figure}

The reduction of the slit opening can be found analogously to Ref.~\cite{Johannes21} (table 3) due to the applied assumption of a single-wall interface. However, the phase shift needs to be adapted and reads for a membrane of thickness $t$ with a slit opening $d$
\begin{eqnarray}
    \varphi(x) \approx - \frac{m\lambda_{\rm dB}}{2\pi\hbar^2} \int U(x,z) \md z= -\frac{12 C_3 m\lambda_{\rm dB} t d \left(d^2+12 x^2\right)}{\hbar^2 \pi \left(d+2x\right)^3\left(d-2x\right)^3}\,. 
\end{eqnarray}
To this end, the transmission function for a single slit adapts due to the dispersion force interaction to
\begin{eqnarray}
    t(x) =\Theta\left(\frac{d-2\Delta R}{2}-x\right)\Theta\left(\frac{d-2\Delta R}{2}+x\right) \mathrm e^{\mi \varphi(x)}\,. \label{eq:transmw}
\end{eqnarray}

The impact of the dispersion forces (Casimir--Polder)  is illustrated in Fig.~\ref{fig:comparison}, which shows the difference in the interference pattern from the classical double slit experiment using a scalar wave and a matter-wave with dispersion force interaction. A much higher population of the higher diffraction orders can be observed.

\subsubsection{Generation of training data}\label{sec:gen_training_data}
In this proof of concept paper we have chosen to consider a  1D system with dimensions so that the Fraunhofer approximation applies. We use a typical wavelength for metastable helium $\lambda_{\rm dB} = 0.1 \,\rm{nm}$, assuming a helium point-source located $L_1 = 1 \,\rm{m}$ away from the mask, and a substrate patterning plane at $L_2 = 300\,\rm{\mu m}$ behind the mask. The mask is considered to be made of SiN with a thickness of $5\,{\rm{nm}}$. The maximum extension of the mask is restricted to $200\,\rm{nm}$ to stay within the Fraunhofer approximation (satisfying a Fresnel number below 1). Furthermore the mask is separated into 50 sections of $4\,\rm{nm}$ width each, corresponding to the minimum opening width. This width was selected based on Ref.~\cite{Johannes21}, which shows that the effective entrance width is reduced by about 2~~nm due to the dispersion force interaction.

We randomly generate masks represented via arrays with entries 0  for close or 1 for open sections. Neighbouring open sections will be combined to a single opening with larger thickness. Hence, the randomly generated array will be transferred into a set of width and position for each opening. Thus, the allowed openings varies between 1 and 50 units. To this end, we calculate the single slit diffraction [integrant in Eq.~(\ref{eq:wave})]
\begin{equation}\label{eq:data_generation}
    \int \mathrm d x t(x;d_n) \mathrm e^{\mi \frac{k_0}{L_2}r x} \,,
\end{equation}
for all possible thicknesses $d_n$ for matter waves~(\ref{eq:transmw}) and electromagnetic waves~(\ref{eq:emw}) via standard numerical integration techniques. These results are tabled and a second program calculates the superposition of multislits to generate the training data.

%{\color{red} Let us discuss this comment, perhaps it should be moved to the discussion: This restriction to Fraunhofer diffraction strongly limits the spatial resolution of the created patterns (50 values for the masks and 1000 points for the interference patterns) To reduce the resolution on the nanometre scale, the masks needs to be extended to be micrometre-sized. Such an extension yields data sets with million of points, which is doable, but requires enormous computational power.}

\subsection{An application example: the double slit pattern}
To demonstrate the performance of \lacenet, we  invert a well-known diffraction pattern: the pattern resulting from a double slit mask, see Fig. \ref{fig:comparison}. Note that a double slit is very different from the \lacenet's training set, which consist of randomly generated masks with no particular preset structure.

\begin{figure}
\centering
\begin{minipage}{.49\textwidth}
  \centering
  \includegraphics[width=\linewidth]{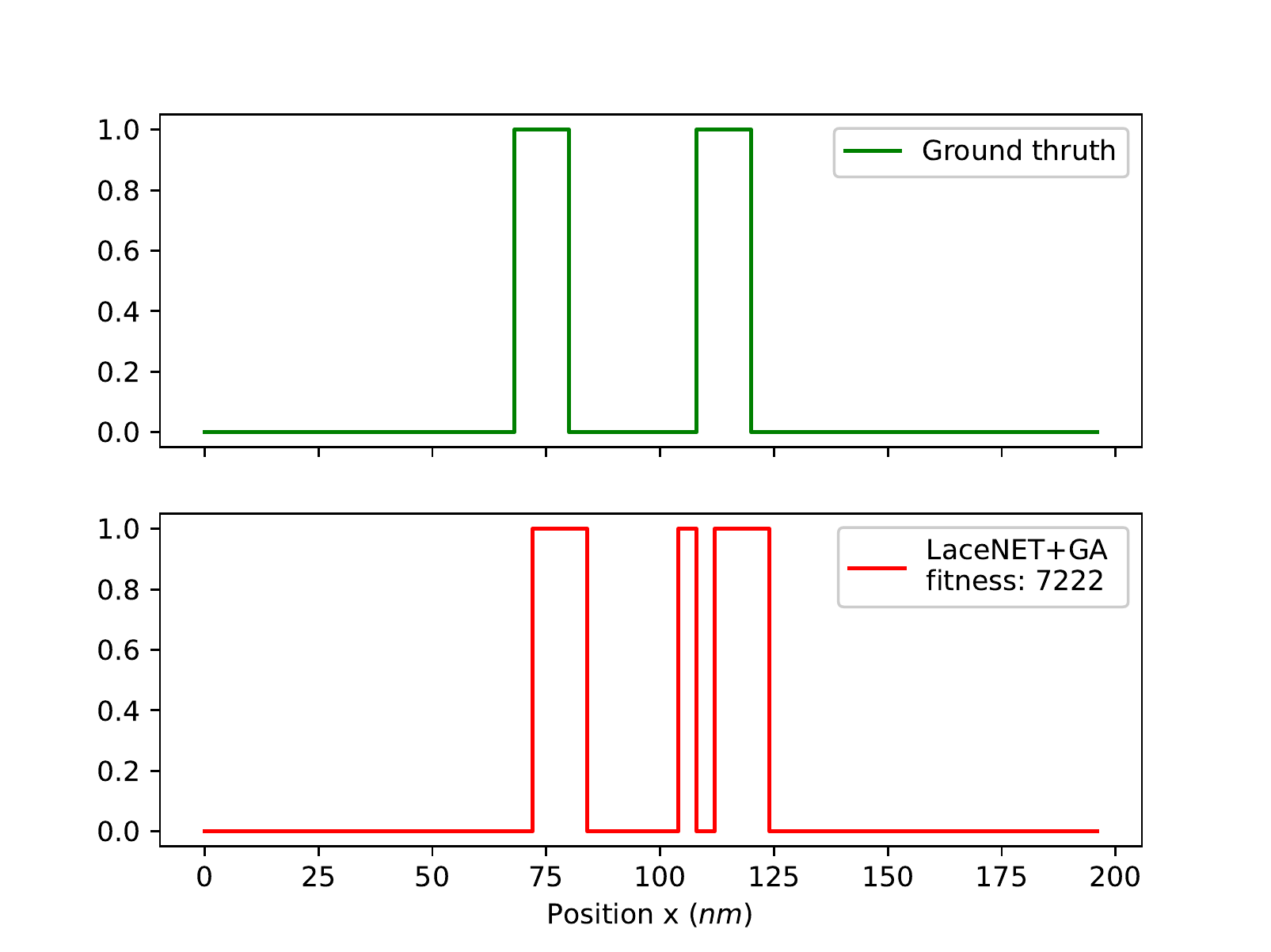}
  \captionof{figure}{Mask returned by \lacenet after attempting to invert the pattern for a metastable helium beam corresponding to a double slit mask with slit width 12 nm, spacing 28 nm (measured between the end of the first slit and the beginning of the second), and an assumed membrane material of 5 nm SiN. Note how the mask is slightly different. However, the pattern produced by it is very similar t the desired one.}
  \label{fig:Lacenet_doubleslit}
\end{minipage}%
\hfill
\begin{minipage}{.49\textwidth}
  \centering
  \includegraphics[width=\linewidth]{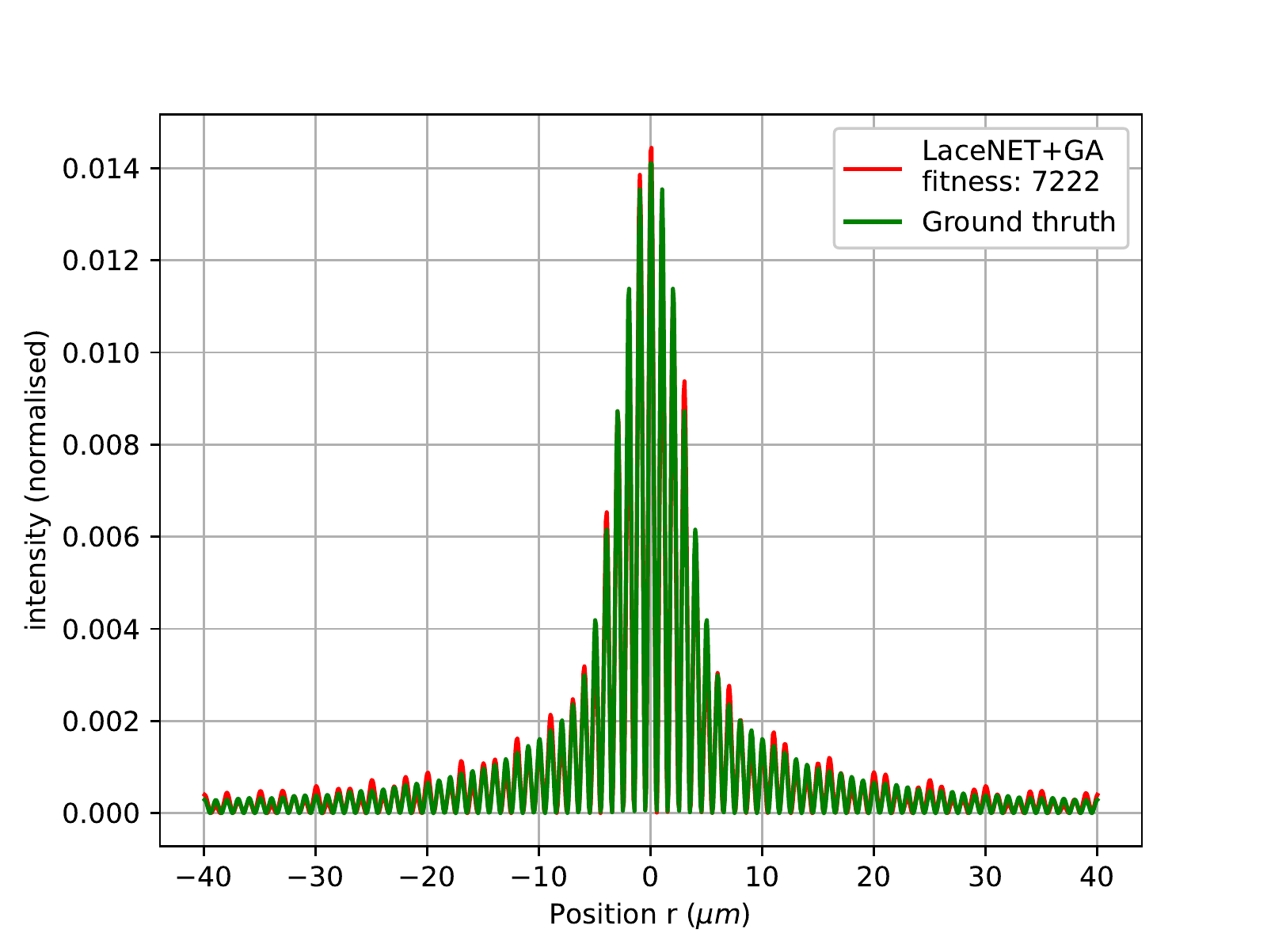}
  \captionof{figure}{Pattern returned by the mask produced by \lacenet after attempting to solve the double slit inversion problem. Note the near-perfect match with the ground truth despite the significant difference in the returned mask (Fig. \ref{fig:Lacenet_doubleslit}).}
  \label{fig:Lacenet_doubleslit_pattern}
\end{minipage}
\end{figure}

\lacenet successfully inverts the pattern to a good degree of accuracy after 10000 generations of the Genetic Algorithm. Interestingly, it does not obtain the exact same mask as the ground truth but the mask that it produces matches very closely to the desired pattern (see Figs. \ref{fig:Lacenet_doubleslit} and \ref{fig:Lacenet_doubleslit_pattern}).

%\begin{figure}
%    \centering
%    \includegraphics[width=0.8\linewidth]{figure_masks_doubleslit.eps}
%    \caption{\label{fig:Lacenet_doubleslit}  Mask returned by \lacenet after attempting to invert the pattern for a metastable helium beam corresponding to a double slit mask with slit width 12 nm, spacing 28 nm (measured between the end of the first slit and the beginning of the second), and an assumed membrane material of 5 nm SiN. Note how the mask is slightly different. However, the pattern produced by it is very similar t the desired one.}
%\end{figure}

%\begin{figure}
%    \centering
%    \includegraphics[width=0.8\linewidth]{figure_intensities_doubleslit.eps}
%    \caption{\label{fig:Lacenet_doubleslit_pattern}  Pattern returned by the mask produced by \lacenet after attempting to solve the double slit inversion problem. Note the near-perfect match with the ground truth despite the significant difference in the returned mask (Fig. \ref{fig:Lacenet_doubleslit}).}
%\end{figure}

\section{Discussion}\label{sec12}
Despite the known capabilities of deep learning as a powerful inverse strategy, we found that the best strategy to obtain accurate result was to couple it with an additional optimisation step (in our case, a Genetic Algorithm). \lacenet's difficulty in fully solving the inverse could be due to (i) insufficient data, (ii) insufficient network complexity, or (iii) the very hard inverse problem that it is trying to solve.

We performed several experiments aimed to determine the cause of \lacenet's difficulty with some examples. Interestingly, increasing the complexity of the network (by adding layers, and increasing the number of neurons in the layers), did not in general produce a better generalisation. With regards to point (ii), we did see a clear improvement with increasing the size of our data-set (which we can arbitrarily scale up by solving for Eq. (\ref{eq:data_generation})). The experiments presented here use a data set of 300k examples. By using the data generation procedure described in Sec. \ref{sec:gen_training_data}, one can extend the data set arbitrarily, potentially increasing the generalisation capabilities of \lacenet.

Finally, we saw some examples for which initialising the mask using a random sequence or using \lacenet made no significant difference. By observing these examples, it seems to be the case that this happens more often for masks that are containing many small openings. These masks are more likely to be automatically randomly produced by the Genetic Algorithm. Therefore, it is possible that for some cases the random initialisation is as good as, or even better than \lacenet's guess.

\section{Conclusion}\label{sec13}
We present a machine learning architecture for solving the inverse problem for matter wave lithography: \lacenet. The architecture of \lacenet consists of a convolutional deep neural network followed by a Genetic Algorithm.  This is a key technological step for industrial scale matter-wave lithography, needed to achieve the ultimate goal of fast patterning of individual atoms and molecules over large areas. The dispersion force interaction for matter waves presents a great computational challenge compared to scalar waves due to the complex integral and its nonlinear solution.

In this proof of concept paper we have restricted ourselves to the 1D case in the Fraunhofer approximation regime. The focus of future work will be to modify \lacenet to solve the general problem in 2D. 

\section*{Acknowledgements}
The computations were performed on the Norwegian Research and Education
Cloud (NREC), using resources provided by the University of
Bergen. http://www.nrec.no/
Special thanks to Tore Burheim and Tor L{\ae}dre 
This work has received funding from the European Union’s Horizon 2020 research and innovation
programme. B.H. gratefully acknowledges support from the  H2020-FETOPEN-2018-2019-2020-01 under Grant Agreement No. 863127 Nanolace (www.nanolace.eu). J.F. gratefully acknowledges support from H2020-MSCA-IF-2020, Grant Agreement No. 101031712.

\appendix
\section{Hyperparameter tuning}\label{secA1}
We optimise our hyperparameters using the Tree-structured Parzen Estimator algorithm \cite{bergstra2011algorithms} - a bayesian algorithm commonly found in hyperparameter optimisation packages. To do so, we use optuna \cite{akiba2019optuna}, an open source hyperparameter search library for python. To find our hyperparameters we do two searches: one in which we focus in finding the best batch size for our problem - which the hyper-parameter optimiser finds to be 225 and another one in which we focus on optimising the parameters of the focal loss within a more constrained space. The complete list of trials and their reported Mean Squared Error can be found in Table \ref{tab:tab1}.
\begin{longtable}{lllllll}
\caption{Results of the hyperparameter tuning rounds in \lacenet \label{tab:tab1}}\\
{} & optimizer &        lr & batch\_size &     alpha &     gamma &     MSE \\ \hline\endfirsthead
\caption{Continuation}\\
{} & optimizer &        lr & batch\_size &     alpha &     gamma &     MSE\\
\hline
\endhead
11 &      Adam &  0.000163 &        225 &  0.439672 &  5.951962 &  1.399777 \\
43 &   RMSprop &  0.000288 &        225 &  0.448443 &  6.431573 &  1.404423 \\
47 &   RMSprop &  0.000123 &         87 &  0.439803 &  6.291578 &  1.421799 \\
48 &   RMSprop &  0.000283 &        200 &  0.448358 &  6.074838 &  1.433143 \\
39 &   RMSprop &  0.000099 &         66 &  0.466304 &  7.131291 &  1.450584 \\
4  &      Adam &  0.000098 &        225 &  0.584123 &  5.253094 &  1.456233 \\
22 &      Adam &  0.000173 &        225 &   0.40145 &  5.910984 &  1.456791 \\
23 &      Adam &  0.000109 &        225 &  0.450265 &  5.525971 &  1.465228 \\
12 &      Adam &  0.000187 &        225 &  0.412308 &   6.12187 &  1.465743 \\
32 &   RMSprop &  0.000089 &        114 &  0.179152 &  7.315897 &  1.466564 \\
40 &   RMSprop &  0.000082 &        570 &  0.471317 &  3.003355 &  1.469744 \\
46 &   RMSprop &  0.000059 &        153 &  0.295671 &  7.276354 &  1.472631 \\
16 &      Adam &  0.000133 &        225 &  0.521078 &   5.03229 &  1.477234 \\
6  &      Adam &  0.000097 &        225 &  0.467451 &   5.42483 &  1.478946 \\
35 &   RMSprop &  0.000179 &         66 &  0.540199 &  7.791814 &  1.485383 \\
2  &   RMSprop &  0.000052 &        225 &  0.487136 &  7.756889 &  1.485913 \\
0  &   RMSprop &  0.000075 &        225 &  0.505513 &  6.664503 &  1.488749 \\
15 &      Adam &  0.000035 &        225 &  0.436437 &  5.701631 &  1.490682 \\
37 &   RMSprop &  0.000151 &         73 &  0.560933 &  6.776026 &  1.495198 \\
28 &   RMSprop &  0.000035 &        265 &  0.364326 &  4.117896 &  1.497929 \\
19 &      Adam &  0.000144 &        225 &  0.582637 &  6.224196 &  1.499658 \\
21 &      Adam &  0.000261 &        225 &  0.422763 &  6.114556 &  1.502561 \\
20 &      Adam &  0.000025 &        225 &  0.401408 &  6.951192 &  1.506756 \\
7  &      Adam &  0.000069 &        225 &  0.496068 &  7.374401 &   1.51577 \\
13 &      Adam &  0.000031 &        225 &  0.452886 &  5.889216 &  1.530536 \\
49 &   RMSprop &  0.000265 &        195 &   0.52036 &   6.28119 &  1.544203 \\
18 &      Adam &  0.000016 &        225 &   0.43794 &   5.56616 &  1.549521 \\
38 &   RMSprop &  0.000434 &        605 &  0.494833 &  6.954664 &  1.549902 \\
29 &      Adam &  0.000039 &        400 &  0.360762 &  7.630682 &  1.560381 \\
45 &   RMSprop &  0.000329 &        332 &  0.413058 &  4.932252 &  1.580241 \\
1  &   RMSprop &  0.000285 &        225 &  0.555782 &  7.176087 &  1.583419 \\
26 &      Adam &  0.000047 &        987 &   0.32565 &   4.25768 &  1.593861 \\
33 &      Adam &  0.000032 &        155 &  0.120995 &  5.757826 &  1.606285 \\
36 &   RMSprop &   0.00018 &        142 &   0.59511 &  7.965855 &  1.611265 \\
27 &   RMSprop &   0.00003 &        391 &  0.234176 &  1.175896 &  1.626354 \\
34 &   RMSprop &  0.000013 &        789 &   0.38773 &   4.12605 &  1.627522 \\
31 &      Adam &  0.000076 &        900 &  0.167727 &  5.418842 &  1.630291 \\
10 &      Adam &  0.000013 &        225 &  0.593005 &  6.110873 &  1.638141 \\
41 &   RMSprop &  0.000013 &        241 &  0.270598 &  5.188218 &  1.647733 \\
25 &      Adam &   0.00029 &        784 &  0.230935 &  2.162501 &  1.686709 \\
8  &   RMSprop &   0.00037 &        225 &  0.478398 &  5.746003 &  1.688195 \\
14 &      Adam &  0.000184 &        225 &  0.533729 &  6.477521 &   1.69213 \\
42 &   RMSprop &  0.000112 &        435 &  0.101017 &  6.930886 &   1.71168 \\
30 &   RMSprop &  0.000706 &        285 &  0.407261 &  6.188819 &  1.725915 \\
24 &      Adam &  0.000514 &        225 &  0.400875 &  5.872189 &  1.773344 \\
5  &   RMSprop &  0.000757 &        225 &  0.415991 &  5.231926 &  1.793177 \\
44 &   RMSprop &  0.000935 &        226 &  0.460966 &  6.326114 &  1.799258 \\
17 &      Adam &   0.00037 &        225 &  0.553661 &  6.577556 &  1.904586 \\
3  &      Adam &  0.000923 &        225 &  0.498035 &   5.16874 &  2.041645 \\
9  &      Adam &  0.000786 &        225 &   0.57071 &  5.326067 &  2.277016 
\end{longtable}
\newpage
\section{Extended results}\label{secA2}
In Figure~\ref{fig:random_vs_lacenet}, we show a randomly selected example from the 100 examples that are averaged in Fig. \ref{fig:deltapercentage}. In this section we show all other examples. There is no inherent physical difference between the figures shown here and  Fig. \ref{fig:deltapercentage} - they simply correspond to different randomly-generated masks. The figures are vectorised, so they can be zoomed in digitally and explored at full resolution. It is evident from the figures presented here that most of them follow the same trend as Fig. \ref{fig:random_vs_lacenet}. 
\foreach \x in {figure_netvsrandom_0,
figure_netvsrandom_1,
figure_netvsrandom_2,
figure_netvsrandom_3}{%
    \includegraphics[width=0.249\linewidth]{\x.pdf}%
}%

\foreach \x in {
figure_netvsrandom_4,
figure_netvsrandom_5,
figure_netvsrandom_6,
figure_netvsrandom_7}{%
    \includegraphics[width=0.249\linewidth]{\x.pdf}%
}%

\foreach \x in {figure_netvsrandom_8,
figure_netvsrandom_9,
figure_netvsrandom_10,
figure_netvsrandom_11}{%
    \includegraphics[width=0.249\linewidth]{\x.pdf}%
}%

\foreach \x in {
figure_netvsrandom_12,
figure_netvsrandom_13,
figure_netvsrandom_14,
figure_netvsrandom_15}{%
    \includegraphics[width=0.249\linewidth]{\x.pdf}%
}%

\foreach \x in {
figure_netvsrandom_16,
figure_netvsrandom_17,
figure_netvsrandom_18,
figure_netvsrandom_19}{%
    \includegraphics[width=0.249\linewidth]{\x.pdf}%
}%

\foreach \x in {
figure_netvsrandom_20,
figure_netvsrandom_21,
figure_netvsrandom_22,
figure_netvsrandom_23}{%
    \includegraphics[width=0.249\linewidth]{\x.pdf}%
}%

\foreach \x in {
figure_netvsrandom_24,
figure_netvsrandom_25,
figure_netvsrandom_26,
figure_netvsrandom_27}{%
    \includegraphics[width=0.249\linewidth]{\x.pdf}%
}%

\foreach \x in {
figure_netvsrandom_28,
figure_netvsrandom_29,
figure_netvsrandom_30,
figure_netvsrandom_31}{%
    \includegraphics[width=0.249\linewidth]{\x.pdf}%
}%

\foreach \x in {
figure_netvsrandom_32,
figure_netvsrandom_33,
figure_netvsrandom_34,
figure_netvsrandom_35}{%
    \includegraphics[width=0.249\linewidth]{\x.pdf}%
}%

\foreach \x in {
figure_netvsrandom_36,
figure_netvsrandom_37,
figure_netvsrandom_38,
figure_netvsrandom_39}{%
    \includegraphics[width=0.249\linewidth]{\x.pdf}%
}%

\foreach \x in {
figure_netvsrandom_40,
figure_netvsrandom_41,
figure_netvsrandom_42,
figure_netvsrandom_43}{%
    \includegraphics[width=0.249\linewidth]{\x.pdf}%
}%

\foreach \x in {
figure_netvsrandom_44,
figure_netvsrandom_45,
figure_netvsrandom_46,
figure_netvsrandom_47}{%
    \includegraphics[width=0.249\linewidth]{\x.pdf}%
}%

\foreach \x in {
figure_netvsrandom_48,
figure_netvsrandom_49,
figure_netvsrandom_50,
figure_netvsrandom_51}{%
    \includegraphics[width=0.249\linewidth]{\x.pdf}%
}%

\foreach \x in {
figure_netvsrandom_52,
figure_netvsrandom_53,
figure_netvsrandom_54,
figure_netvsrandom_55}{%
    \includegraphics[width=0.249\linewidth]{\x.pdf}%
}%

\foreach \x in {
figure_netvsrandom_56,
figure_netvsrandom_57,
figure_netvsrandom_58,
figure_netvsrandom_59}{%
    \includegraphics[width=0.249\linewidth]{\x.pdf}%
}%

\foreach \x in {
figure_netvsrandom_60,
figure_netvsrandom_61,
figure_netvsrandom_62,
figure_netvsrandom_63}{%
    \includegraphics[width=0.249\linewidth]{\x.pdf}%
}%

\foreach \x in {
figure_netvsrandom_64,
figure_netvsrandom_65,
figure_netvsrandom_66,
figure_netvsrandom_67}{%
    \includegraphics[width=0.249\linewidth]{\x.pdf}%
}%

\foreach \x in {
figure_netvsrandom_68,
figure_netvsrandom_69,
figure_netvsrandom_70,
figure_netvsrandom_71}{%
    \includegraphics[width=0.249\linewidth]{\x.pdf}%
}%

\foreach \x in {
figure_netvsrandom_72,
figure_netvsrandom_73,
figure_netvsrandom_74,
figure_netvsrandom_75}{%
    \includegraphics[width=0.249\linewidth]{\x.pdf}%
}%

\foreach \x in {
figure_netvsrandom_76,
figure_netvsrandom_77,
figure_netvsrandom_78,
figure_netvsrandom_79}{%
    \includegraphics[width=0.249\linewidth]{\x.pdf}%
}%

\foreach \x in {figure_netvsrandom_80,
figure_netvsrandom_81,
figure_netvsrandom_82,
figure_netvsrandom_83}{%
    \includegraphics[width=0.249\linewidth]{\x.pdf}%
}%

\foreach \x in {
figure_netvsrandom_84,
figure_netvsrandom_85,
figure_netvsrandom_86,
figure_netvsrandom_87}{%
    \includegraphics[width=0.249\linewidth]{\x.pdf}%
}%

\foreach \x in {figure_netvsrandom_88,
figure_netvsrandom_89,
figure_netvsrandom_90,
figure_netvsrandom_91}{%
    \includegraphics[width=0.249\linewidth]{\x.pdf}%
}%

\foreach \x in {
figure_netvsrandom_92,
figure_netvsrandom_93,
figure_netvsrandom_94,
figure_netvsrandom_95}{%
    \includegraphics[width=0.249\linewidth]{\x.pdf}%
}%

\foreach \x in {
figure_netvsrandom_96,
figure_netvsrandom_97,
figure_netvsrandom_98,
figure_netvsrandom_99}{%
    \includegraphics[width=0.249\linewidth]{\x.pdf}%
}%

%Bibliography
\bibliographystyle{unsrt}  
\bibliography{sn-bibliography}

\end{document}